\pdfoutput=1
\documentclass[11pt]{article}

\usepackage[final]{acl}

\usepackage{times}
\usepackage{latexsym}

\usepackage[T1]{fontenc}
\usepackage[utf8]{inputenc}

\usepackage{microtype}

\usepackage{inconsolata}

\usepackage{graphicx}
\usepackage{fontawesome5}
\usepackage{float}
\usepackage{placeins}
\usepackage{float}
\usepackage{multirow}
\usepackage{xcolor}
\usepackage{ulem}
\usepackage{booktabs}
\usepackage{adjustbox}
\usepackage{amssymb}
\usepackage{array} 
\usepackage{rotating}
\usepackage{tabularray}
\usepackage{subcaption}   % For creating subfigures
\usepackage{float} 
\usepackage{amsmath}
\usepackage{float}

\title{LLM-Mixer: Multiscale Mixing in LLMs for Time Series Forecasting}

\author{
  \textbf{Md Kowsher\textsuperscript{1}},
  \textbf{Md. Shohanur Islam Sobuj\textsuperscript{2}},
  \textbf{Nusrat Jahan Prottasha\textsuperscript{1}},
\\
  \textbf{E. Alejandro Alanis\textsuperscript{3}},
  \textbf{Ozlem Ozmen Garibay\textsuperscript{1}},
  \textbf{Niloofar Yousefi\textsuperscript{1}}
\\
\\
  \textsuperscript{1}University of Central Florida, USA,
  \textsuperscript{2}Anymate Me, Germany,
  \textsuperscript{3}Microsoft, USA
\\
\faGithub~\href{https://github.com/Kowsher/LLMMixer}{\textcolor{red}{\texttt{https://github.com/Kowsher/LLMMixer}}}
  \small{
  }
}

\begin{document}
\maketitle
\begin{abstract}
Time series forecasting is a challenging task, especially when dealing with data that contains both short-term variations and long-term trends. In this study, we introduce LLM-Mixer, a novel framework that combines multiscale time-series decomposition with the power of pre-trained Large Language Models (LLMs). LLM-Mixer breaks down time-series data into multiple temporal resolutions using downsampling and processes these multiscale representations with a frozen LLM, guided by a carefully designed text prompt that encodes information about the dataset's features and structure. To understand the role of downsampling, we conduct a detailed analysis using Neural Tangent Kernel (NTK) distance, showing that incorporating multiple scales improves the model’s learning dynamics.
We evaluate LLM-Mixer across a diverse set of forecasting tasks, including long-term multivariate, short-term multivariate, and long-term univariate scenarios. Experimental results demonstrate that LLM-Mixer achieves competitive performance compared to recent state-of-the-art models across various forecasting horizons. Code is available at: \href{https://github.com/Kowsher/LLMMixer}{https://github.com/Kowsher/LLMMixer}
\end{abstract}

\section{Introduction \& Related Work}

Time series forecasting is essential in numerous fields, including finance \cite{zhang2024deep}, energy management \cite{martin2010prediction}, healthcare \cite{morid2023time}, climate science \cite{mudelsee2019trend}, and industrial operations \cite{wang2020industrial}. Traditional forecasting models, such as AutoRegressive Integrated Moving Average (ARIMA) \cite{box2015time} and exponential smoothing techniques \cite{hyndman2018forecasting}, are widely used for straightforward predictive tasks. However, these models assume stationarity and linearity, which limit their effectiveness when applied to complex, nonlinear, and multivariate time series often found in real-world scenarios \cite{cheng2015time}. The advent of deep learning has significantly advanced time series forecasting. CNNs \cite{wang2023micn, tang2020rethinking, kirisci2022new} have been utilized for capturing temporal patterns, while RNNs \cite{siami2019performance, zhang2019lstm, karim2019insights} are adept at modeling temporal state transitions. However, both CNNs and RNNs have limitations in capturing long-term dependencies \cite{wang2024timemixer, tang2021building, zhu2023long}. Recently, Transformer architectures \cite{vaswani2017attention} have demonstrated strong capabilities in handling both local and long-range dependencies, making them suitable for time series forecasting \cite{liu2024itransformer, nie2022time, woo2022etsformer}.

\begin{figure*}[!htb]
%\begin{wrapfigure}{r}{0.7\textwidth}
%\captionsetup{font=footnotesize}

 \begin{center}

      \includegraphics[width=0.9941\linewidth]{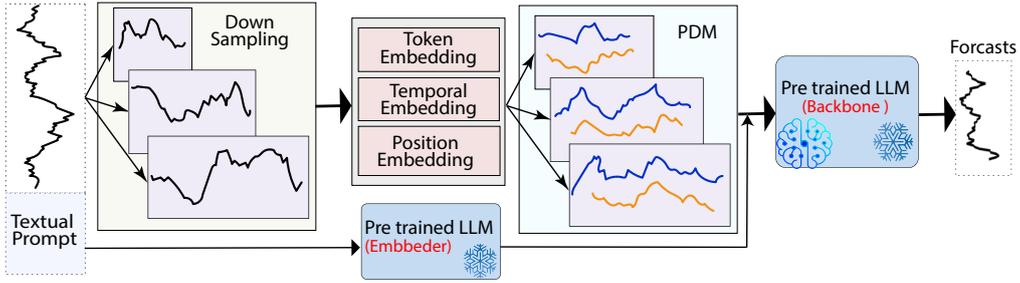}
   \end{center}
%\end{wrapfigure}
\caption{The LLM-Mixer framework for time series forecasting. Time series data is downsampled to multiple scales and enriched with embeddings. These multiscale representations are processed by the Past-Decomposable-Mixing (PDM) module and then input into a pre-trained LLM, which, guided by a textual description, generates the forecast.}
\label{fig:method}
\end{figure*}

In parallel, pre-trained LLMs such as GPT-3 \cite{brown2020language}, GPT-4 \cite{achiam2023gpt}, and LLaMA \cite{touvron2023llama} have achieved remarkable generalization in natural language processing tasks \cite{friha2024llm} due to capabilities of few-shot or zero-shot transfer learning \cite{brown2020language}, multimodal knowledge \cite{jia2024llm} and reasoning \cite{liu2024much}. These models are now being applied across various fields, including computer vision \cite{bendou2024llm}, healthcare \cite{gebreab2024llm},  and finance \cite{zhao2024revolutionizing}. Recently, a few studies have explored using LLMs for time series forecasting due to their impressive capabilities \cite{jin2024timellm, jin2023time, gruver2023large}. However, adapting LLMs to time series data presents challenges because there are significant differences between token-based text data and continuous time series data \cite{morales2024developing}. LLMs are built to handle discrete tokens, which limits their ability to capture the continuous and often irregular patterns found in time series data. Additionally, time series data has multiple time scales, from short-term fluctuations to long-term trends, making it difficult for traditional LLMs to capture all these patterns at once. LLMs typically process fixed-length sequences, which means they may only capture short-term dependencies if the sequence length (i.e., the window of time steps) is small. However, extending the sequence length to capture long-term trends increases computational costs and may dilute the model’s ability to focus on short-term fluctuations within the same sequence. Previous studies using LLMs on time series data have mostly fed the original or a single sequence directly into a frozen LLM, making it hard for the model to fully understand these sequences \cite{jin2024timellm, jin2023time, gruver2023large}.

To address this, we introduce \textbf{LLM-Mixer}, which breaks down the time series data into multiple time scales. By creating various resolutions (Figure~\ref{fig:method}), our model can capture both short-term details and long-term patterns more effectively. Since the LLM remains frozen during training, the multiscale decomposition provides a diverse range of temporal information, helping the model better understand complex time series data.

 \textbf{Our contributions} of this paper are:
\textbf{(1)} We propose \textbf{LLM-Mixer}, a new method that adapts LLMs for time series forecasting by breaking down the data into different time scales, helping the model capture both short-term and long-term patterns. \textbf{(2)} Our method creates multiple versions of the time series at different resolutions which helps the LLM to understand complex time series data more effectively. \textbf{(3)} Empirical results show that \textbf{LLM-Mixer} achieves competitive performance, improves forecasting accuracy on both multivariate and univariate data, and works effectively for both short-term and long-term forecasting tasks.

\section{LLM Mixer}
\textbf{Preliminaries:} In multivariate time series forecasting, we are given historical data \(\mathbf{X} = \{ \mathbf{x}_1, \dots, \mathbf{x}_T \} \in \mathbb{R}^{T \times M}\), where \(T\) is the number of time steps and \(M\) is the number of features. The goal is to predict the future values for the next \(K\) time steps, denoted as \(\mathbf{Y} = \{ \mathbf{x}_{T+1}, \dots, \mathbf{x}_{T+K} \} \in \mathbb{R}^{K \times M}\). For convenience, let \(\mathbf{X}_{t,:}\) represent the data at time step \(t\), and \(\mathbf{X}_{:,m}\) represent the full time series for variable \(m \in M\). 

Now, suppose we have a prompt \(\mathbf{P}\), which includes textual information about the time sequence (e.g., source, features, distribution, statistics). We use a pre-trained language model \(\mathbb{F}(\cdot)\) with frozen parameters \(\Theta\), then the prediction is made as follows:
\[
\hat{\mathbf{Y}} = \mathbb{F}(\mathbf{X}, \mathbf{P}; \Theta, \mathbf{\Phi})
\]
Here \(\mathbf{\Phi}\) is a small set of trainable parameters to adjust the model for the specific forecasting task.

\noindent\textbf{Multi-scale View of Time Data:}
Time series data contains patterns at various levels—small scales capture detailed changes, while larger scales highlight overarching trends \cite{liu2022optoelectronic, mozer1991induction}. Analyzing data at multiple scales helps to understand these complex patterns \cite{wang2024timemixer}. Following \cite{wang2024timemixer}, we apply a multiscale mixing strategy. First, we downsample the time series \(\mathbf{X}\) into \(\tau\) scales using average pooling, resulting in a multiscale representation \(\mathbf{\mathcal{X}} = \{ \mathbf{x}_0, \mathbf{x}_1, \dots, \mathbf{x}_\tau \}\), where each \(\mathbf{x}_i \in \mathbb{R}^{\frac{T}{2^i} \times M}\). Here, \(\mathbf{x}_0\) contains the finest temporal details, while \(\mathbf{x}_\tau\) captures the broadest trends. 

Next, we project these multiscale series into deep features using three types of embeddings: token, temporal, and positional embeddings. Token embeddings are obtained via 1D convolutions \cite{kiranyaz20211d}, temporal embeddings represent day, week, and month \cite{jimenez2023embedded}, and positional embeddings encode sequence positions.

We then use stacked Past-Decomposable-Mixing (PDM) blocks by following the framework from~\cite{wang2024timemixer, jimenez2023embedded}  to mix past information across different scales. PDM  works by breaking down complex time series data into separate seasonal and trend components at multiple scales, allowing for targeted processing of each component by using the framework from \cite{wang2024timemixer, wu2021autoformer}.  For the \(l\)-th layer, PDM is defined as 
\vspace{-0.050cm}
\[\mathbf{\mathcal{X}}^l = \textit{PDM} (\mathbf{\mathcal{X}}^{l-1}), \quad l \in L\]
where \(L\) is the total number of layers, and \(\mathbf{\mathcal{X}}^l = \{ \mathbf{x}_0^l, \mathbf{x}_1^l, \dots, \mathbf{x}_\tau^l \}\), with each \(\mathbf{x}_i^l \in \mathbb{R}^{\frac{T}{2^i} \times d}\), where \(d\) is the model's dimension.
\begin{table*}[!htb]

\resizebox{\textwidth}{!}{%
\begin{tabular}{cc|cc|cc|cc|ll|ll|ll|cc|cc|cc|ll|ll|ll}
\hline
\multicolumn{2}{c|}{Methods} & \multicolumn{2}{c|}{\begin{tabular}[c]{@{}c@{}}LLM-Mixer\\ (llama2)\end{tabular}} & \multicolumn{2}{c|}{\begin{tabular}[c]{@{}c@{}}LLM-Mixer\\ (roberta)\end{tabular}} & \multicolumn{2}{c|}{TIME-LLM} & \multicolumn{2}{c|}{TimeMixer} & \multicolumn{2}{c|}{iTransformer} & \multicolumn{2}{c|}{RLinear} & \multicolumn{2}{c|}{DLinear} & \multicolumn{2}{c|}{PatchTST} & \multicolumn{2}{c|}{TimesNet} & \multicolumn{2}{c|}{TiDE} & \multicolumn{2}{c|}{TimesNet} & \multicolumn{2}{c}{Crossformer} \\ \hline
\multicolumn{2}{c|}{Metric} & MSE & MAE & MSE & MAE & MSE & MAE & \multicolumn{1}{c}{MSE} & \multicolumn{1}{c|}{MAE} & \multicolumn{1}{c}{MSE} & \multicolumn{1}{c|}{MAE} & \multicolumn{1}{c}{MSE} & \multicolumn{1}{c|}{MAE} & MSE & MAE & MSE & MAE & MSE & MAE & \multicolumn{1}{c}{MSE} & \multicolumn{1}{c|}{MAE} & \multicolumn{1}{c}{MSE} & \multicolumn{1}{c|}{MAE} & \multicolumn{1}{c}{MSE} & \multicolumn{1}{c}{MAE} \\ \hline

\multicolumn{1}{c|}{\multirow{5}{*}{ETTh1}} & 96 & 
\textcolor{red}{\textbf{0.368}} & \textcolor{red}{\textbf{0.395}} & 0.372 & 0.399 & \textcolor{blue}{\uline{0.369}} & \textcolor{blue}{\uline{0.397}} & 0.375 & 0.400 & 0.386 & 0.405 & 0.386 & 0.395 & 0.397 & 0.412 & 0.460 & 0.447 & 0.384 & 0.402 & 0.479 & 0.464 & 0.384 & 0.402 & 0.423 & 0.448 \\

\multicolumn{1}{c|}{} & 192
    & \textcolor{red}{\textbf{0.406}} & \textcolor{red}{\textbf{0.417}} & 0.439 & 0.470 & \textcolor{blue}{\uline{0.411}} & 0.428 & 0.429 & 0.421 & 0.441 & 0.436 & 0.437 & \textcolor{blue}{\uline{0.424}} & 0.446 & 0.441 & 0.512 & 0.477 & 0.436 & 0.429 & 0.525 & 0.492 & 0.436 & 0.429 & 0.471 & 0.474 \\

\multicolumn{1}{c|}{} & 336 & \textcolor{blue}{\uline{0.446}} & \textcolor{red}{\textbf{0.444}} & 0.458 & 0.467 & \textcolor{red}{\textbf{0.440}} & 0.447 & 0.484 & 0.458 & 0.487 & 0.458 & 0.479 & \textcolor{blue}{\uline{0.446}} & 0.489 & 0.467 & 0.546 & 0.496 & 0.638 & 0.469 & 0.565 & 0.515 & 0.491 & 0.469 & 0.570 & 0.546 \\
\multicolumn{1}{c|}{} & 720 & \textcolor{red}{\textbf{0.461}} & \textcolor{red}{\textbf{0.475}} & 0.465 & 0.480 & \textcolor{blue}{\uline{0.462}} & \textcolor{blue}{\uline{0.477}} & 0.498 & 0.482 & 0.503 & 0.491 & 0.481 & 0.470 & 0.513 & 0.510 & 0.544 & 0.517 & 0.521 & 0.500 & 0.594 & 0.558 & 0.521 & 0.500 & 0.653 & 0.621 \\ \cline{2-26}

\multicolumn{1}{c|}{} & Avg & \textcolor{red}{\textbf{0.420}} & \textcolor{red}{\textbf{0.433}} & 0.434 & 0.454 & \textcolor{blue}{\uline{0.421}} & 0.437 & 0.447 & 0.440 & 0.454 & 0.447 & 0.446 & \textcolor{blue}{\uline{0.434}} & 0.461 & 0.457 & 0.516 & 0.484 & 0.495 & 0.450 & 0.541 & 0.507 & 0.458 & 0.450 & 0.529 & 0.522 \\ \hline

\multicolumn{1}{c|}{\multirow{5}{*}{ETTh2}} & 96 & \textcolor{red}{\textbf{0.274}} & \textcolor{red}{\textbf{0.334}} & 0.284 & 0.347 & \textcolor{blue}{\uline{0.278}} & \textcolor{blue}{\uline{0.338}} & 0.289 & 0.341 & 0.297 & 0.349 & 0.288 & 0.338 & 0.340 & 0.394 & 0.308 & 0.355 & 0.340 & 0.374 & 0.400 & 0.440 & 0.340 & 0.374 & 0.745 & 0.584 \\

\multicolumn{1}{c|}{} & 192 & \textcolor{blue}{\uline{0.339}} & 0.384 & 0.343 & \textcolor{blue}{\uline{0.389}} & \textcolor{red}{\textbf{0.338}} & \textcolor{red}{\textbf{0.384}} & 0.372 & 0.392 & 0.380 & 0.400 & 0.374 & 0.390 & 0.482 & 0.479 & 0.393 & 0.405 & 0.402 & 0.414 & 0.528 & 0.509 & 0.402 & 0.414 & 0.877 & 0.656 \\

\multicolumn{1}{c|}{} & 336 & \textcolor{blue}{\uline{0.380}} & \textcolor{red}{\textbf{0.408}} & \textcolor{red}{\textbf{0.375}} & \textcolor{blue}{\uline{0.409}} & 0.389 & 0.411 & 0.386 & 0.414 & 0.428 & 0.432 & 0.415 & 0.426 & 0.591 & 0.541 & 0.427 & 0.436 & 0.452 & 0.452 & 0.643 & 0.571 & 0.452 & 0.452 & 1.043 & 0.731 \\

\multicolumn{1}{c|}{} & 720 & \textcolor{red}{\textbf{0.390}} & \textcolor{red}{\textbf{0.431}} & 0.394 & 0.438 & \textcolor{blue}{\uline{0.393}} & \textcolor{blue}{\uline{0.432}} & 0.412 & 0.434 & 0.427 & 0.445 & 0.420 & 0.440 & 0.839 & 0.661 & 0.436 & 0.450 & 0.462 & 0.468 & 0.874 & 0.679 & 0.462 & 0.468 & 1.104 & 0.763 \\ \cline{2-26}

\multicolumn{1}{c|}{} & Avg & \textcolor{red}{\textbf{0.345}} & \textcolor{red}{\textbf{0.389}} & 0.349 & 0.395 & \textcolor{blue}{\uline{0.349}} & \textcolor{blue}{\uline{0.391}} & 0.364 & 0.395 & 0.383 & 0.407 & 0.374 & 0.398 & 0.563 & 0.519 & 0.391 & 0.411 & 0.414 & 0.427 & 0.611 & 0.550 & 0.414 & 0.427 & 0.942 & 0.684 \\ \hline

\multicolumn{1}{c|}{\multirow{5}{*}{ETTm1}} & 96 & \textcolor{blue}{\uline{0.294}} & \textcolor{blue}{\uline{0.346}} & 0.304 & 0.348 & \textcolor{red}{\textbf{0.293}} & \textcolor{red}{\textbf{0.343}} & 0.320 & 0.357 & 0.334 & 0.368 & 0.355 & 0.376 & 0.346 & 0.374 & 0.352 & 0.374 & 0.338 & 0.375 & 0.364 & 0.387 & 0.338 & 0.375 & 0.404 & 0.426 \\

\multicolumn{1}{c|}{} & 192 & \textcolor{red}{\textbf{0.348}} & \textcolor{red}{\textbf{0.367}} & 0.350 & 0.377 & \textcolor{blue}{\uline{0.350}} & \textcolor{blue}{\uline{0.368}} & 0.361 & 0.381 & 0.377 & 0.391 & 0.391 & 0.392 & 0.382 & 0.391 & 0.390 & 0.393 & 0.374 & 0.387 & 0.398 & 0.404 & 0.374 & 0.387 & 0.450 & 0.451 \\

\multicolumn{1}{c|}{} & 336 & \textcolor{blue}{\uline{0.387}} & \textcolor{blue}{\uline{0.392}} & 0.395 & 0.409 & \textcolor{red}{\textbf{0.382}} & \textcolor{red}{\textbf{0.391}} & 0.390 & 0.404 & 0.426 & 0.420 & 0.424 & 0.415 & 0.415 & 0.415 & 0.421 & 0.414 & 0.410 & 0.411 & 0.428 & 0.425 & 0.410 & 0.411 & 0.532 & 0.515 \\

\multicolumn{1}{c|}{} & 720 & \textcolor{blue}{\uline{0.439}} & \textcolor{blue}{\uline{0.442}} & 0.448 & 0.450 & \textcolor{red}{\textbf{0.443}} & 0.451 & 0.454 & \textcolor{red}{\textbf{0.441}} & 0.491 & 0.459 & 0.487 & 0.450 & 0.473 & 0.451 & 0.462 & 0.449 & 0.478 & 0.450 & 0.487 & 0.461 & 0.478 & 0.450 & 0.666 & 0.589 \\ \cline{2-26}

\multicolumn{1}{c|}{} & Avg &  \textcolor{red}{\textbf{0.367}} &  \textcolor{red}{\textbf{0.387}} & 0.374 & 0.396 & 0.367 & \textcolor{blue}{\uline{0.388}} & \textcolor{blue}{\uline{0.381}} & 0.395 & 0.407 & 0.410 & 0.414 & 0.407 & 0.404 & 0.408 & 0.406 & 0.407 & 0.400 & 0.406 & 0.419 & 0.419 & 0.400 & 0.406 & 0.513 & 0.495 \\ \hline

\multicolumn{1}{c|}{\multirow{5}{*}{ETTm2}} & 96 & \textcolor{red}{\textbf{0.160}} & \textcolor{red}{\textbf{0.251}} & 0.160 & 0.253 & 0.160 & 0.251 & \textcolor{blue}{\uline{0.175}} & \textcolor{blue}{\uline{0.252}} & 0.180 & 0.264 & 0.182 & 0.265 & 0.193 & 0.293 & 0.183 & 0.270 & 0.187 & 0.267 & 0.207 & 0.305 & 0.187 & 0.267 & 0.287 & 0.366 \\

\multicolumn{1}{c|}{} & 192 & \textcolor{blue}{\uline{0.226}} & \textcolor{red}{\textbf{0.290}} & 0.229 & 0.297 & \textcolor{red}{\textbf{0.220}} & \textcolor{blue}{\uline{0.292}} & 0.237 & 0.299 & 0.250 & 0.309 & 0.246 & 0.304 & 0.284 & 0.361 & 0.255 & 0.314 & 0.249 & 0.309 & 0.290 & 0.364 & 0.249 & 0.309 & 0.414 & 0.492 \\

\multicolumn{1}{c|}{} & 336 & \textcolor{red}{\textbf{0.283}} & \textcolor{blue}{\uline{0.339}} & 0.299 & 0.346 & \textcolor{blue}{\uline{0.284}} & \textcolor{red}{\textbf{0.337}} & 0.298 & 0.340 & 0.311 & 0.348 & 0.307 & 0.342 & 0.382 & 0.429 & 0.309 & 0.347 & 0.321 & 0.351 & 0.377 & 0.422 & 0.321 & 0.351 & 0.597 & 0.542 \\

\multicolumn{1}{c|}{} & 720 & \textcolor{blue}{\uline{0.392}} & 0.398 & 0.399 & 0.405 & 0.391 & \textcolor{blue}{\uline{0.397}}
 & \textcolor{red}{\textbf{0.391}} & \textcolor{red}{\textbf{0.396}} & 0.412 & 0.407 & 0.407 & 0.398 & 0.558 & 0.525 & 0.412 & 0.404 & 0.365 & 0.359 & 0.558 & 0.524 & 0.408 & 0.403 & 1.730 & 1.042 \\ \cline{2-26}

\multicolumn{1}{c|}{} & Avg & \textcolor{blue}{\uline{0.265}} & \textcolor{blue}{\uline{0.320}} & 0.272 & 0.325 & \textcolor{red}{\textbf{0.264}} & \textcolor{red}{\textbf{0.319}} & 0.275 & 0.323 & 0.288 & 0.332 & 0.286 & 0.327 & 0.354 & 0.402 & 0.290 & 0.334 & 0.291 & 0.333 & 0.358 & 0.404 & 0.291 & 0.333 & 0.757 & 0.610 \\ \hline

\multicolumn{1}{c|}{\multirow{5}{*}{Weather}} & 96 & \textcolor{blue}{\uline{0.149}} & 0.202 & 0.151 & \textcolor{blue}{\uline{0.203}} & \textcolor{red}{\textbf{0.148}} & \textcolor{red}{\textbf{0.202}} & 0.163 & 0.209 & 0.174 & 0.214 & 0.192 & 0.232 & 0.195 & 0.252 & 0.186 & 0.227 & 0.172 & 0.220 & 0.202 & 0.261 & 0.172 & 0.220 & 0.195 & 0.271 \\

\multicolumn{1}{c|}{} & 192 & \textcolor{red}{\textbf{0.197}} & \textcolor{red}{\textbf{0.239}} & 0.209 & 0.249 & \textcolor{blue}{\uline{0.199}} & \textcolor{blue}{\uline{0.242}} & 0.208 & 0.250 & 0.221 & 0.254 & 0.240 & 0.271 & 0.237 & 0.295 & 0.234 & 0.265 & 0.219 & 0.261 & 0.242 & 0.298 & 0.219 & 0.261 & 0.209 & 0.277 \\

\multicolumn{1}{c|}{} & 336 & 0.270 & 0.282 & 0.310 & \textcolor{blue}{\uline{0.281}} & 0.262 & \textcolor{red}{\textbf{0.279}} & \textcolor{blue}{\uline{0.251}} & 0.287 & 0.278 & 0.296 & 0.292 & 0.307 & 0.282 & 0.331 & 0.284 & 0.301 & \textcolor{red}{\textbf{0.246}} & 0.337 & 0.287 & 0.335 & 0.280 & 0.306 & 0.273 & 0.332 \\

\multicolumn{1}{c|}{} & 720 & 0.323 & 0.332 & 0.339 & 0.342 & 0.330 & 0.334 & 0.339 & 0.341 & 0.358 & 0.347 & 0.364 & 0.353 & \textcolor{blue}{\uline{0.282}} & \textcolor{blue}{\uline{0.331}} & 0.356 & 0.349 & 0.365 & 0.359 & 0.287 & 0.335 & \textcolor{red}{\textbf{0.280}} & \textcolor{red}{\textbf{0.306}} & 0.379 & 0.401 \\ \cline{2-26}

\multicolumn{1}{c|}{} & Avg & \textcolor{red}{\textbf{0.235}} & \textcolor{red}{\textbf{0.264}} & 0.252 & \textcolor{blue}{\uline{0.269}} & 0.235 & 0.264 & \textcolor{blue}{\uline{0.240}} & 0.271 & 0.258 & 0.278 & 0.272 & 0.291 & 0.265 & 0.315 & 0.265 & 0.285 & 0.251 & 0.294 & 0.271 & 0.320 & 0.259 & 0.287 & 0.264 & 0.320 \\ \hline

\multicolumn{1}{c|}{\multirow{5}{*}{Electricity}} & 96 & \textcolor{blue}{\uline{0.143}} & \textcolor{red}{\textbf{0.233}} & 0.150 & 0.241 & \textcolor{red}{\textbf{0.142}} & \textcolor{blue}{\uline{0.234}} & 0.153 & 0.247 & 0.148 & 0.240 & 0.201 & 0.281 & 0.210 & 0.302 & 0.190 & 0.296 & 0.168 & 0.272 & 0.237 & 0.329 & 0.168 & 0.272 & 0.219 & 0.314 \\

\multicolumn{1}{c|}{} & 192 & \textcolor{red}{\textbf{0.151}} & \textcolor{blue}{\uline{0.242}} & 0.166 & 0.259 & \textcolor{blue}{\uline{0.152}} & \textcolor{red}{\textbf{0.241}} & 0.166 & 0.256 & 0.162 & 0.253 & 0.201 & 0.283 & 0.210 & 0.305 & 0.199 & 0.304 & 0.184 & 0.322 & 0.236 & 0.330 & 0.184 & 0.289 & 0.231 & 0.322 \\

\multicolumn{1}{c|}{} & 336 & \textcolor{red}{\textbf{0.178}} & \textcolor{blue}{\uline{0.267}} & \textcolor{blue}{\uline{0.180}} & 0.281 & 0.180 & \textcolor{red}{\textbf{0.263}} & 0.185 & 0.277 & 0.178 & 0.269 & 0.215 & 0.298 & 0.223 & 0.319 & 0.217 & 0.319 & 0.198 & 0.300 & 0.249 & 0.344 & 0.198 & 0.300 & 0.246 & 0.337 \\

\multicolumn{1}{c|}{} & 720 & \textcolor{red}{\textbf{0.213}} & \textcolor{red}{\textbf{0.305}} & 0.221 & 0.311 & \textcolor{blue}{\uline{0.218}} & \textcolor{blue}{\uline{0.308}} & 0.225 & 0.310 & 0.225 & 0.317 & 0.257 & 0.331 & 0.258 & 0.350 & 0.258 & 0.352 & 0.220 & 0.320 & 0.284 & 0.373 & 0.220 & 0.320 & 0.280 & 0.363 \\ \cline{2-26}

\multicolumn{1}{c|}{} & Avg & \textcolor{red}{\textbf{0.171}} & \textcolor{red}{\textbf{0.253}} & 0.174 & 0.273 & \textcolor{blue}{\uline{0.173}} & \textcolor{blue}{\uline{0.261}} & 0.182 & 0.272 & 0.178 & 0.270 & 0.219 & 0.298 & 0.225 & 0.319 & 0.216 & 0.318 & 0.193 & 0.304 & 0.251 & 0.344 & 0.192 & 0.295 & 0.244 & 0.334 \\ \hline

\multicolumn{1}{c|}{\multirow{5}{*}{Traffic}} & 96 & \textcolor{red}{\textbf{0.380}} & \textcolor{red}{\textbf{0.264}} & 0.394 & 0.274 & \textcolor{blue}{\uline{0.382}} & \textcolor{blue}{\uline{0.268}} & 0.462 & 0.285 & 0.395 & 0.268 & 0.649 & 0.389 & 0.650 & 0.396 & 0.526 & 0.347 & 0.593 & 0.321 & 0.805 & 0.493 & 0.593 & 0.321 & 0.644 & 0.429 \\

\multicolumn{1}{c|}{} & 192 & \textcolor{blue}{\uline{0.396}} & \textcolor{blue}{\uline{0.269}} & 0.399 & 0.276 &  \textcolor{red}{\textbf{0.394}} &  \textcolor{red}{\textbf{0.267}} & 0.473 & 0.296 & 0.417 & 0.276 & 0.601 & 0.366 & 0.598 & 0.370 & 0.522 & 0.332 & 0.617 & 0.336 & 0.756 & 0.474 & 0.617 & 0.336 & 0.665 & 0.431 \\

\multicolumn{1}{c|}{} & 336 & \textcolor{red}{\textbf{0.423}} & \textcolor{red}{\textbf{0.274}} & 0.439 & \textcolor{blue}{\uline{0.280}} & \textcolor{blue}{\uline{0.425}} & 0.281 & 0.498 & 0.296 & 0.433 & 0.283 & 0.609 & 0.369 & 0.605 & 0.373 & 0.517 & 0.334 & 0.629 & 0.336 & 0.762 & 0.477 & 0.629 & 0.336 & 0.674 & 0.420 \\

\multicolumn{1}{c|}{} & 720 & \textcolor{red}{\textbf{0.458}} & \textcolor{red}{\textbf{0.296}} & \textcolor{blue}{\uline{0.460}} & \textcolor{blue}{\uline{0.298}} & 0.460 & 0.300 & 0.506 & 0.313 & 0.467 & 0.302 & 0.647 & 0.387 & 0.645 & 0.394 & 0.552 & 0.352 & 0.640 & 0.350 & 0.719 & 0.449 & 0.640 & 0.350 & 0.683 & 0.424 \\ \cline{2-26}

\multicolumn{1}{c|}{} & Avg & \textcolor{red}{\textbf{0.414}} & \textcolor{red}{\textbf{0.265}} & 0.433 & 0.282 & \textcolor{blue}{\uline{0.415}} & \textcolor{blue}{\uline{0.279}} & 0.484 & 0.297 & 0.428 & 0.282 & 0.626 & 0.378 & 0.625 & 0.383 & 0.529 & 0.341 & 0.620 & 0.336 & 0.760 & 0.473 & 0.620 & 0.336 & 0.667 & 0.426 \\ \hline
\end{tabular}%
}
\caption{Full long-term multivariate forecasting results. \textcolor{red}{\textbf{Red:}} the best, \textcolor{blue}{\uline{Blue}}: the second best.} \label{tab:full_mul}
\end{table*}

\begin{table*}[!htb]
% updated

\resizebox{\textwidth}{!}{%
\begin{tabular}{cc|ll|ll|ll|ll|cc|cc|cc|cc|cc|cc|cc|cc}
\hline
\multicolumn{2}{c|}{Methods} & \multicolumn{2}{c|}{\begin{tabular}[c]{@{}c@{}}LLM-Mixer\\ (llama2)\end{tabular}} & \multicolumn{2}{c|}{\begin{tabular}[c]{@{}c@{}}LLM-Mixer\\ (roberta)\end{tabular}} & \multicolumn{2}{c|}{TIME-LLM} & \multicolumn{2}{c|}{TimeMixer} & \multicolumn{2}{c|}{iTransformer} & \multicolumn{2}{c|}{RLinear} & \multicolumn{2}{c|}{PatchTST} & \multicolumn{2}{c|}{Crossformer} & \multicolumn{2}{c|}{TiDE} & \multicolumn{2}{c|}{TimesNet} & \multicolumn{2}{c|}{DLinear} & \multicolumn{2}{c}{SCINet} \\ \hline
\multicolumn{2}{c|}{Metric} & \multicolumn{1}{c}{MSE} & \multicolumn{1}{c|}{MAE} & \multicolumn{1}{c}{MSE} & \multicolumn{1}{c|}{MAE} & \multicolumn{1}{c}{MSE} & \multicolumn{1}{c|}{MAE} & \multicolumn{1}{c}{MSE} & \multicolumn{1}{c|}{MAE} & MSE & MAE & MSE & MAE & MSE & MAE & MSE & MAE & MSE & MAE & MSE & MAE & MSE & MAE & MSE & MAE \\ \hline
\multicolumn{1}{c|}{\multirow{5}{*}{PEMS03}} 
& 12 
& \textcolor{blue}{\uline{0.069}} & \textcolor{blue}{\uline{0.173}} 
& 0.082 & 0.190 
& 0.092 & 0.201 
& 0.082 & 0.189 
& 0.071 & 0.174 
& 0.126 & 0.236 
& 0.099 & 0.216 
& 0.090 & 0.203 
& 0.178 & 0.305 
& 0.085 & 0.192 
& 0.122 & 0.243 
& \textcolor{red}{\textbf{0.066}} & \textcolor{red}{\textbf{0.172}} \\

\multicolumn{1}{c|}{} 
& 24 
& \textcolor{blue}{\uline{0.090}} & 0.200
& 0.092 & 0.201 
& 0.095 & 0.207 
& 0.090 & \textcolor{blue}{\uline{0.199}} 
& 0.093 & 0.201 
& 0.246 & 0.334 
& 0.142 & 0.259 
& 0.121 & 0.240 
& 0.257 & 0.371 
& 0.118 & 0.223 
& 0.201 & 0.317 
& \textcolor{red}{\textbf{0.085}} & \textcolor{red}{\textbf{0.198}} \\

\multicolumn{1}{c|}{} 
& 48 
& \textcolor{red}{\textbf{0.123}} & \textcolor{red}{\textbf{0.232}} 
& 0.126 & 0.237 
& 0.127 & 0.237 
& 0.125 & 0.235 
& \textcolor{blue}{\uline{0.125}} & \textcolor{blue}{\uline{0.236}} 
& 0.551 & 0.529 
& 0.211 & 0.319 
& 0.202 & 0.317 
& 0.379 & 0.463 
& 0.155 & 0.260 
& 0.333 & 0.425 
& 0.127 & 0.238 \\

\multicolumn{1}{c|}{} & 96 & 0.165 & \textcolor{red}{\textbf{0.274}} & 0.166 & 0.276 & 0.165 & 0.274 & 0.167 & 0.275 & \textcolor{blue}{\uline{0.164}} & \textcolor{blue}{\uline{0.275}} & 1.057 & 0.787 & 0.269 & 0.370 & 0.262 & 0.367 & 0.490 & 0.539 & \textcolor{red}{\textbf{0.028}} & 0.317 & 0.457 & 0.515 & 0.178 & 0.287 \\ \cline{2-26} 
\multicolumn{1}{c|}{} & Avg & \textcolor{red}{\textbf{0.112}} & \textcolor{red}{\textbf{0.220}} & 0.116 & 0.226 & 0.120 & 0.230 & 0.116 & 0.225 & \textcolor{blue}{\uline{0.113}} & \textcolor{blue}{\uline{0.221}} & 0.495 & 0.472 & 0.180 & 0.291 & 0.169 & 0.281 & 0.326 & 0.419 & 0.147 & 0.248 & 0.278 & 0.375 & 0.114 & 0.224 \\ \hline

\multicolumn{1}{c|}{\multirow{5}{*}{PEMS04}} 
& 12 & \textcolor{blue}{\uline{0.072}} & \textcolor{red}{\textbf{0.177}} & 0.076 & 0.183 & \textcolor{red}{\textbf{0.071}} & 0.177 & 0.073 & \textcolor{blue}{\uline{0.179}} & 0.078 & 0.183 & 0.138 & 0.252 & 0.105 & 0.224 & 0.098 & 0.209 & 0.182 & 0.324 & 0.087 & 0.195 & 0.148 & 0.272 & 0.073 & 0.177 \\

\multicolumn{1}{c|}{} 
& 24 & \textcolor{blue}{\uline{0.095}} & \textcolor{blue}{\uline{0.201}} & 0.105 & 0.211 & 0.106 & 0.214 & 0.097 & 0.205 & 0.108 & 0.205 & 0.258 & 0.387 & 0.150 & 0.266 & 0.135 & 0.250 & 0.309 & 0.454 & 0.099 & 0.217 & 0.225 & 0.367 & \textcolor{red}{\textbf{0.084}} & \textcolor{red}{\textbf{0.193}} \\

\multicolumn{1}{c|}{} 
& 48 & \textcolor{red}{\textbf{0.099}} & \textcolor{red}{\textbf{0.216}} & 0.103 & 0.220 & \textcolor{blue}{\uline{0.101}} & 0.218 & 0.099 & 0.217 & 0.120 & 0.233 & 0.572 & 0.544 & 0.229 & 0.339 & 0.205 & 0.353 & 0.470 & 0.539 & 0.135 & 0.253 & 0.355 & 0.437 & 0.099 & \textcolor{blue}{\uline{0.217}} \\

\multicolumn{1}{c|}{} 
& 96 & \textcolor{blue}{\uline{0.120}} & \textcolor{red}{\textbf{0.225}} & 0.130 & \textcolor{blue}{\uline{0.232}} & 0.121 & 0.225 & 0.121 & 0.225 & 0.150 & 0.267 & 1.159 & 0.947 & 0.309 & 0.520 & 0.299 & 0.467 & 0.656 & 0.637 & \textcolor{red}{\textbf{0.043}} & 0.317 & 0.550 & 0.541 & 0.129 & \textcolor{blue}{\uline{0.227}} \\ \cline{2-26}

\multicolumn{1}{c|}{} 
& Avg & \textcolor{red}{\textbf{0.097}} & \textcolor{red}{\textbf{0.205}} & 0.104 & 0.211 & 0.100 & 0.209 & \textcolor{blue}{\uline{0.098}} & \textcolor{blue}{\uline{0.207}} & 0.111 & 0.221 & 0.526 & 0.491 & 0.195 & 0.307 & 0.209 & 0.314 & 0.353 & 0.475 & 0.129 & 0.245 & 0.329 & 0.395 & 0.119 & 0.234 \\ \hline

\multicolumn{1}{c|}{\multirow{5}{*}{PEMS07}} 
& 12 & \textcolor{red}{\textbf{0.065}} & \textcolor{red}{\textbf{0.165}} & 0.072 & 0.180 & 0.068 & \textcolor{blue}{\uline{0.166}} & 0.070 & 0.168 & \textcolor{blue}{\uline{0.067}} & 0.165 & 0.118 & 0.235 & 0.097 & 0.226 & 0.093 & 0.209 & 0.155 & 0.324 & 0.081 & 0.185 & 0.118 & 0.272 & 0.068 & 0.174 \\

\multicolumn{1}{c|}{} 
& 24 
& \textcolor{red}{\textbf{0.087}} & \textcolor{red}{\textbf{0.105}} 
& \textcolor{blue}{\uline{0.091}} & 0.198 
& 0.088 & 0.192 
& 0.087 & 0.105 
& 0.088 & 0.190
& 0.271 & 0.449 
& 0.153 & 0.276 
& 0.138 & 0.251 
& 0.338 & 0.475 
& 0.096 & 0.223 
& 0.207 & 0.381 
& 0.087 & \textcolor{blue}{\uline{0.180}} \\

\multicolumn{1}{c|}{} 
& 48 
& \textcolor{red}{\textbf{0.106}} & \textcolor{red}{\textbf{0.215}} 
& 0.117 & 0.224 
& 0.109 & 0.219 
& 0.106 & \textcolor{blue}{\uline{0.217}} 
& \textcolor{blue}{\uline{0.110}} & 0.215 
& 0.596 & 0.621 
& 0.343 & 0.459 
& 0.309 & 0.401 
& 0.532 & 0.547 
& 0.145 & 0.264 
& 0.593 & 0.484 
& 0.149 & 0.233 \\

\multicolumn{1}{c|}{} 
& 96 
& 0.147 & 0.266 
& \textcolor{blue}{\uline{0.146}} & \textcolor{blue}{\uline{0.265}} 
& 0.150 & 0.269 
& 0.151 & 0.269 
& \textcolor{red}{\textbf{0.141}} & \textcolor{red}{\textbf{0.245}} 
& 1.096 & 0.795 
& 0.346 & 0.490 
& 0.329 & 0.443 
& 0.674 & 0.650 
& 0.203 & 0.307 
& 0.789 & 0.531 
& 0.191 & 0.267 \\

\cline{2-26} 
\multicolumn{1}{c|}{} 
& Avg 
& \textcolor{red}{\textbf{0.101}} & \textcolor{red}{\textbf{0.204}} 
& 0.107 & 0.217 
& \textcolor{blue}{\uline{0.104}} & \textcolor{blue}{\uline{0.212}}
& \textcolor{blue}{\uline{0.104}} & 0.218 
& 0.101 & 0.204 
& 0.504 & 0.478 
& 0.213 & 0.303 
& 0.215 & 0.326 
& 0.355 & 0.499 
& 0.129 & 0.245 
& 0.529 & 0.387 
& 0.119 & 0.234 \\ \hline

\multicolumn{1}{c|}{\multirow{5}{*}{PEMS08}} 
& 12 
& 0.082 & 0.186 
& 0.086 & 0.190 
& \textcolor{blue}{\uline{0.080}} & 0.184 
& 0.083 & \textcolor{blue}{\uline{0.183}}
& \textcolor{red}{\textbf{0.079}} & \textcolor{red}{\textbf{0.182}} 
& 0.133 & 0.247 
& 0.168 & 0.232 
& 0.152 & 0.267 
& 0.215 & 0.367 
& 0.154 & 0.276 
& 0.172 & 0.291 
& 0.087 & 0.184 \\

\multicolumn{1}{c|}{} 
& 24 
& 0.107 & \textcolor{blue}{\uline{0.213}} 
& 0.109 & 0.217 
& \textcolor{blue}{\uline{0.105}} & \textcolor{blue}{\uline{0.208}} 
& 0.109 & 0.218 
& 0.105 & 0.209 
& 0.242 & 0.360 
& 0.189 & 0.321 
& 0.174 & 0.314 
& 0.258 & 0.430 
& 0.178 & 0.307 
& 0.290 & 0.346 
& \textcolor{red}{\textbf{0.104}} & \textcolor{red}{\textbf{0.193}} \\

\multicolumn{1}{c|}{} 
& 48 
& 0.187 & \textcolor{blue}{\uline{0.235}}
& 0.192 & 0.240 
& 0.193 & 0.242 
& 0.192 & 0.241 
& \textcolor{blue}{\uline{0.125}} & 0.237 
& 0.596 & 0.556 
& 0.269 & 0.389 
& 0.247 & 0.388 
& 0.433 & 0.512 
& 0.210 & 0.345 
& 0.418 & 0.422 
& \textcolor{red}{\textbf{0.124}} & \textcolor{red}{\textbf{0.216}} \\

\multicolumn{1}{c|}{} 
& 96 
& \textcolor{red}{\textbf{0.140}} & \textcolor{red}{\textbf{0.245}} 
& 0.142 & \textcolor{blue}{\uline{0.249}}
& 0.147 & 0.256 
& 0.143 & 0.252 
& 0.167 & 0.275
& \textcolor{blue}{\uline{1.043}} & 0.841 
& 0.344 & 0.470 
& 0.326 & 0.459 
& 0.534 & 0.571 
& 0.283 & 0.387 
& 0.593 & 0.514 
& 0.155 & 0.253 \\

\cline{2-26} 
\multicolumn{1}{c|}{} 
& Avg 
& \textcolor{blue}{\uline{0.119}} & 0.226 
& 0.132 & 0.224 
& 0.131 & \textcolor{blue}{\uline{0.223}}
& 0.132 & 0.224 
& 0.119 & 0.226 
& 0.503 & 0.501 
& 0.243 & 0.353 
& 0.225 & 0.357 
& 0.360 & 0.470 
& 0.206 & 0.329 
& 0.368 & 0.393 
& \textcolor{red}{\textbf{0.118}} & \textcolor{red}{\textbf{0.212}} \\ \hline

\end{tabular}%
}
\caption{Full short-term multivariate forecasting results. \textcolor{red}{\textbf{Red:}} the best, \textcolor{blue}{\uline{Blue}}: the second best.} \label{tab:full_short}
\end{table*}

\noindent\textbf{Prompt Embedding:} Prompting is an effective technique for guiding LLMs by using task-specific information \cite{sahoo2024systematic,li2023prompt}. Studies like \cite{xue2023promptcast} show promising results by treating time series inputs as prompts for forecasting. \cite{jin2024timellm} further improved time series predictions by embedding dataset descriptions in the prompts. Inspired by this, we embed dataset descriptions (e.g., features, statistics, distribution) as prompts. We use a textual description for all samples in a dataset, as suggested by \cite{jin2024timellm}, and generate its embedding using the pre-trained LLM’s word embeddings, denoted by \(E \in \mathbb{R}^{V \times d}\), where \(V\) is the LLM’s vocabulary size. This prompt leverages the LLM's semantic knowledge to improve the prediction task.

\noindent\textbf{Multi-scale Mixing in LLM:}
After processing through \(L\) PDM blocks, we obtain the multiscale past information \( \mathbf{\mathcal{X}}^L \). Since different scales focus on different variations, their predictions offer complementary strengths. To fully utilize this, we concatenate all the scales and input them into a frozen pre-trained LLM along with the prompt as $\mathbb{F}(E \oplus \mathbf{\mathcal{X}}^L)$. Finally, a trainable decoder (simple linear transformation) with parameters \(\mathbf{\Phi}\) is applied to the last hidden layer of the LLM to predict the next \(K\) future time steps.

\begin{table*}[!htb]

\resizebox{\textwidth}{!}{%
\begin{tabular}{cc|cc|cc|cc|cc|cc|cc|cc|cc|cc|cc}
\hline
\multicolumn{2}{c|}{Methods} & \multicolumn{2}{c|}{\begin{tabular}[c]{@{}c@{}}LLM-Mixer\\ (llama2)\end{tabular}} & \multicolumn{2}{c|}{\begin{tabular}[c]{@{}c@{}}LLM-Mixer\\ (Roberta)\end{tabular}} & \multicolumn{2}{c|}{Linear} & \multicolumn{2}{c|}{NLinear} & \multicolumn{2}{c|}{DLinear} & \multicolumn{2}{c|}{FEDformer-f} & \multicolumn{2}{c|}{FEDformer-w} & \multicolumn{2}{c|}{Autoformer} & \multicolumn{2}{c|}{Informer} & \multicolumn{2}{c}{LogTrans} \\ \hline
\multicolumn{2}{c|}{Metric} & MSE & MAE & MSE & MAE & MSE & MAE & MSE & MAE & MSE & MAE & MSE & MAE & MSE & MAE & MSE & MAE & MSE & MAE & MSE & MAE \\ \hline
\multicolumn{1}{c|}{\multirow{5}{*}{\textit{ETTh1}}} & 96 & \textcolor{red}{\textbf{0.052}} & \textcolor{red}{\textbf{0.175}} & \textcolor{blue}{\uline{0.053}} & 0.177 & 0.189 & 0.359 & 0.055 & \textcolor{blue}{\uline{0.176}} & 0.056 & 0.180 & 0.079 & 0.215 & 0.080 & 0.214 & 0.071 & 0.206 & 0.193 & 0.377 & 0.283 & 0.468 \\
\multicolumn{1}{c|}{} & 192 & \textcolor{red}{\textbf{0.064}} & \textcolor{red}{\textbf{0.200}} & \textcolor{blue}{\uline{0.066}} & \textcolor{blue}{\uline{0.203}} & 0.078 & 0.212 & 0.069 & 0.204 & 0.071 & 0.204 & 0.104 & 0.245 & 0.105 & 0.256 & 0.114 & 0.262 & 0.217 & 0.395 & 0.234 & 0.409 \\
\multicolumn{1}{c|}{} & 336 & \textcolor{red}{\textbf{0.080}} & \textcolor{red}{\textbf{0.226}} & \textcolor{blue}{\uline{0.081}} & \textcolor{blue}{\uline{0.226}} & 0.091 & 0.237 & 0.084 & 0.228 & 0.098 & 0.244 & 0.119 & 0.270 & 0.120 & 0.269 & 0.107 & 0.258 & 0.202 & 0.381 & 0.386 & 0.546 \\
\multicolumn{1}{c|}{} & 720 & \textcolor{red}{\textbf{0.075}} & \textcolor{red}{\textbf{0.222}} & \textcolor{blue}{\uline{0.078}} & \textcolor{blue}{\uline{0.223}} & 0.172 & 0.340 & 0.080 & 0.226 & 0.189 & 0.359 & 0.142 & 0.299 & 0.127 & 0.280 & 0.126 & 0.283 & 0.183 & 0.355 & 0.475 & 0.629 \\
\multicolumn{1}{c|}{} & Avg & \textcolor{red}{\textbf{0.068}} & \textcolor{red}{\textbf{0.206}} & \textcolor{blue}{\uline{0.071}} & \textcolor{blue}{\uline{0.207}} & 0.133 & 0.287 & 0.071 & 0.208 & 0.104 & 0.247 & 0.111 & 0.257 & 0.108 & 0.255 & 0.106 & 0.252 & 0.199 & 0.377 & 0.345 & 0.513 \\ \hline
\multicolumn{1}{c|}{\multirow{5}{*}{\textit{ETTh2}}} & 96 & \textcolor{blue}{\uline{0.125}} & \textcolor{blue}{\uline{0.274}} & \textcolor{red}{\textbf{0.123}} & 0.276 & 0.133 & 0.283 & 0.129 & 0.278 & 0.131 & 0.279 & 0.128 & \textcolor{red}{\textbf{0.271}} & 0.156 & 0.306 & 0.153 & 0.306 & 0.213 & 0.373 & 0.217 & 0.379 \\
\multicolumn{1}{c|}{} & 192 & \textcolor{red}{\textbf{0.166}} & \textcolor{blue}{\uline{0.322}} & \textcolor{blue}{\uline{0.169}} & 0.324 & 0.176 & \textcolor{red}{\textbf{0.330}} & 0.169 & 0.324 & 0.176 & 0.329 & 0.185 & 0.330 & 0.238 & 0.380 & 0.204 & 0.351 & 0.227 & 0.387 & 0.281 & 0.429 \\
\multicolumn{1}{c|}{} & 336 & \textcolor{red}{\textbf{0.193}} & \textcolor{red}{\textbf{0.353}} & \textcolor{blue}{\uline{0.194}} & 0.356 & 0.213 & 0.371 & 0.194 & \textcolor{blue}{\uline{0.355}} & 0.209 & 0.367 & 0.231 & 0.378 & 0.271 & 0.412 & 0.246 & 0.389 & 0.242 & 0.401 & 0.293 & 0.437 \\
\multicolumn{1}{c|}{} & 720 & \textcolor{blue}{\uline{0.222}} & \textcolor{red}{\textbf{0.380}} & 0.225 & \textcolor{blue}{\uline{0.381}} & 0.292 & 0.440 & 0.225 & 0.381 & 0.276 & 0.426 & 0.278 & 0.420 & 0.288 & 0.438 & 0.268 & 0.409 & 0.291 & 0.439 & \textcolor{red}{\textbf{0.218}} & 0.387 \\
\multicolumn{1}{c|}{} & Avg & \textcolor{red}{\textbf{0.177}} & \textcolor{red}{\textbf{0.332}} & \textcolor{blue}{\uline{0.178}} & \textcolor{blue}{\uline{0.334}} & 0.204 & 0.356 & 0.179 & 0.335 & 0.198 & 0.350 & 0.205 & 0.350 & 0.238 & 0.384 & 0.218 & 0.364 & 0.243 & 0.400 & 0.252 & 0.408 \\ \hline
\multicolumn{1}{c|}{\multirow{5}{*}{\textit{ETTm1}}} & 96 & \textcolor{red}{\textbf{0.023}} & \textcolor{red}{\textbf{0.118}} & \textcolor{blue}{\uline{0.026}} & 0.125 & 0.028 & 0.125 & 0.026 & \textcolor{blue}{\uline{0.122}} & 0.028 & 0.123 & 0.033 & 0.140 & 0.036 & 0.149 & 0.056 & 0.183 & 0.109 & 0.277 & 0.049 & 0.171 \\
\multicolumn{1}{c|}{} & 192 & \textcolor{red}{\textbf{0.033}} & \textcolor{red}{\textbf{0.145}} & \textcolor{blue}{\uline{0.036}} & \textcolor{blue}{\uline{0.147}} & 0.043 & 0.154 & 0.039 & 0.149 & 0.045 & 0.156 & 0.058 & 0.186 & 0.069 & 0.206 & 0.081 & 0.216 & 0.151 & 0.310 & 0.157 & 0.317 \\
\multicolumn{1}{c|}{} & 336 & \textcolor{blue}{\uline{0.053}} & \textcolor{red}{\textbf{0.172}} & 0.054 & \textcolor{blue}{\uline{0.176}} & 0.059 & 0.180 & \textcolor{red}{\textbf{0.052}} & 0.172 & 0.061 & 0.182 & 0.084 & 0.231 & 0.071 & 0.209 & 0.076 & 0.218 & 0.427 & 0.591 & 0.289 & 0.459 \\
\multicolumn{1}{c|}{} & 720 & \textcolor{red}{\textbf{0.071}} & \textcolor{blue}{\uline{0.205}} & \textcolor{blue}{\uline{0.072}} & \textcolor{red}{\textbf{0.204}} & 0.080 & 0.211 & 0.073 & 0.207 & 0.080 & 0.210 & 0.102 & 0.250 & 0.105 & 0.248 & 0.110 & 0.267 & 0.438 & 0.586 & 0.430 & 0.579 \\
\multicolumn{1}{c|}{} & Avg & \textcolor{red}{\textbf{0.045}} & \textcolor{red}{\textbf{0.161}} & \textcolor{blue}{\uline{0.047}} & \textcolor{blue}{\uline{0.163}} & 0.053 & 0.167 & 0.048 & 0.163 & 0.054 & 0.168 & 0.069 & 0.202 & 0.070 & 0.203 & 0.081 & 0.221 & 0.281 & 0.441 & 0.231 & 0.381 \\ \hline
\multicolumn{1}{c|}{\multirow{5}{*}{\textit{ETTm2}}} & 96 & \textcolor{red}{\textbf{0.062}} & \textcolor{red}{\textbf{0.180}} & 0.064 & \textcolor{blue}{\uline{0.181}} & 0.066 & 0.189 & \textcolor{blue}{\uline{0.063}} & 0.182 & 0.063 & 0.183 & 0.067 & 0.198 & 0.063 & 0.189 & 0.065 & 0.189 & 0.088 & 0.225 & 0.075 & 0.208 \\
\multicolumn{1}{c|}{} & 192 & \textcolor{blue}{\uline{0.090}} & \textcolor{blue}{\uline{0.222}} & \textcolor{red}{\textbf{0.089}} & \textcolor{red}{\textbf{0.220}} & 0.094 & 0.230 & 0.090 & 0.223 & 0.092 & 0.227 & 0.102 & 0.245 & 0.110 & 0.252 & 0.118 & 0.256 & 0.132 & 0.283 & 0.129 & 0.275 \\
\multicolumn{1}{c|}{} & 336 & \textcolor{red}{\textbf{0.114}} & \textcolor{red}{\textbf{0.255}} & \textcolor{blue}{\uline{0.116}} & \textcolor{blue}{\uline{0.257}} & 0.120 & 0.263 & 0.117 & 0.259 & 0.119 & 0.261 & 0.130 & 0.279 & 0.147 & 0.301 & 0.154 & 0.305 & 0.180 & 0.336 & 0.154 & 0.302 \\
\multicolumn{1}{c|}{} & 720 & \textcolor{blue}{\uline{0.169}} & \textcolor{red}{\textbf{0.313}} & 0.171 & \textcolor{blue}{\uline{0.314}} & 0.175 & 0.320 & 0.170 & 0.318 & 0.175 & 0.320 & 0.178 & 0.325 & 0.219 & 0.368 & 0.182 & 0.335 & 0.300 & 0.435 & \textcolor{red}{\textbf{0.160}} & 0.321 \\
\multicolumn{1}{c|}{} & Avg & \textcolor{red}{\textbf{0.109}} & \textcolor{red}{\textbf{0.243}} & \textcolor{blue}{\uline{0.110}} & \textcolor{blue}{\uline{0.243}} & 0.114 & 0.250 & 0.110 & 0.246 & 0.112 & 0.248 & 0.119 & 0.262 & 0.135 & 0.279 & 0.130 & 0.271 & 0.150 & 0.295 & 0.130 & 0.277 \\ \hline
\end{tabular}%
}
\caption{Full univariate long sequence time-series forecasting results on ETT full benchmark. \textcolor{red}{\textbf{Red:}} the best, \textcolor{blue}{\uline{Blue}}: the second best.} \label{tab:full_uni}
\end{table*}

\section{Experiments}
We evaluate our LLM-Mixer on several datasets commonly used for benchmarking long-term and short-term multivariate forecasting and compared with SOTA baselines. For long-term forecasting, we use the ETT datasets (ETTh1, ETTh2, ETTm1, ETTm2) from \cite{zhou2021informer}, as well as the Weather, Electricity, and Traffic datasets from \cite{zeng2023transformers}. For short-term forecasting, we use the PeMS dataset \cite{chen2001freeway}, which consists of four public traffic network datasets (PEMS03, PEMS04, PEMS07, and PEMS08) with time series collected at various frequencies. We used RoBERTa-base \cite{liu2019roberta} as a medium-sized language model and LLaMA2-7B \cite{touvron2023llama} as a large language model as the backbone of our framework.  

\noindent{\textbf{Baselines}} \label{sec:Baselines}
We compare our model with well-established time-series forecasting baselines  such as TimeMixer \cite{wang2024timemixer}, iTransformer \cite{liu2024itransformer}, TimeLLM \cite{jin2024timellm}, RLinear \cite{li2024revisiting}, SCINet \cite{liu2022scinet}, TimesNet \cite{wu2022timesnet}, TiDE \cite{das2023longterm}, DLinear \cite{zeng2023transformers}, PatchTST \cite{nie2022time}, FEDformer \cite{zhou2022fedformer}, Stationary \cite{liu2022non}, ESTformer \cite{woo2022etsformer}, LightTS \cite{campos2023lightts}, and Autoformer \cite{chen2021autoformer}. Additionally, we include LLM-based systems such as TimeLLM \cite{jin2024timellm} and GPT2TS \cite{zhou2023one}. For multivariate time series forecasting, we follow the setup of \cite{wang2024timemixer}. For short-term forecasting, we adopt the settings from \cite{liu2024itransformer}, and for univariate forecasting, we adhere to the approach in \cite{zeng2023transformers}.

 \noindent\textbf{{Implementation Details}} \label{sec:implementation} All experiments in this work are implemented using PyTorch. We utilize the Hugging Face library for the LLM model. Experiments were conducted on an NVIDIA H100 GPU with 80 GB RAM.

\noindent\textbf{Hyperparameters:} For long-term experiments, a look-back window of 96 is used to predict the next 96 (future context) and 192 (forecast horizons), while short-term experiments use windows of 24 and 48. All experiments run for 10 epochs with a batch size of 64 for RoBERTa and a batch size of 8 with gradient accumulation of 4 for LLaMA2. The ADAM optimizer is employed with default settings \((\beta_1, \beta_2) = (0.9, 0.999)\) and a learning rate of 0.0001. Downsampling levels range from 2 to 5 across all experiments.  For the baseline models, we have followed their original works, with differences only in batch size and learning rate to align with our experimental setup.

\noindent\textbf{Multivariate forecasting results: } LLM-Mixer demonstrates competitive performance in multivariate long forecasting, as shown in Table \ref{tab:full_mul}. Averaged over four forecasting horizons (96, 192, 384, and 720), LLM-Mixer achieves consistently low MSE and MAE values across most datasets, particularly excelling on ETTh1, ETTh2, and Electricity. Compared to other models such as TIME-LLM, TimeMixer, and PatchTST, LLM-Mixer performs favorably, showing that its design effectively captures both short- and long-term dependencies. Notably, LLM-Mixer also exhibits robustness on challenging datasets such as Traffic, where it outperforms several baseline models. These results highlight the efficacy of the LLM-Mixer in handling complex temporal patterns over extended horizons. 

\begin{figure*}[!htb]
    \centering
     \begin{center} \includegraphics[width=0.901\linewidth]{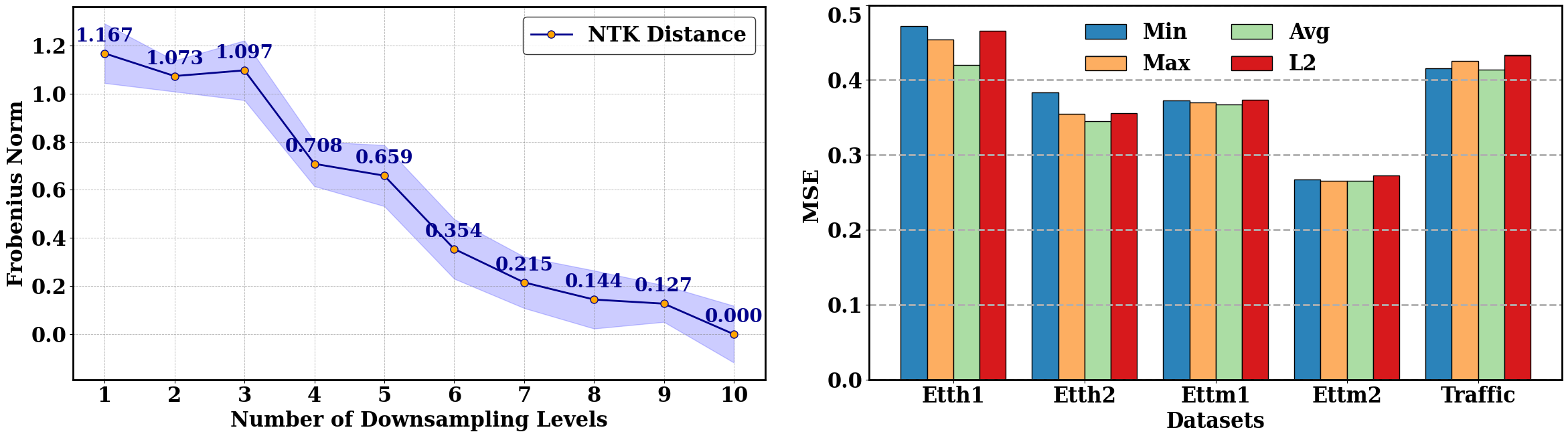}
   \end{center}
    \caption{(\textbf{Left}) Frobenius norm of NTK distance. (\textbf{Right}) Pooling technique for Multi-scale Mixing}
    \label{fig:ablation}
\end{figure*}

\begin{figure*}[!htb]
    \centering
    \begin{subfigure}[b]{\textwidth}
        \includegraphics[width=0.95\textwidth]{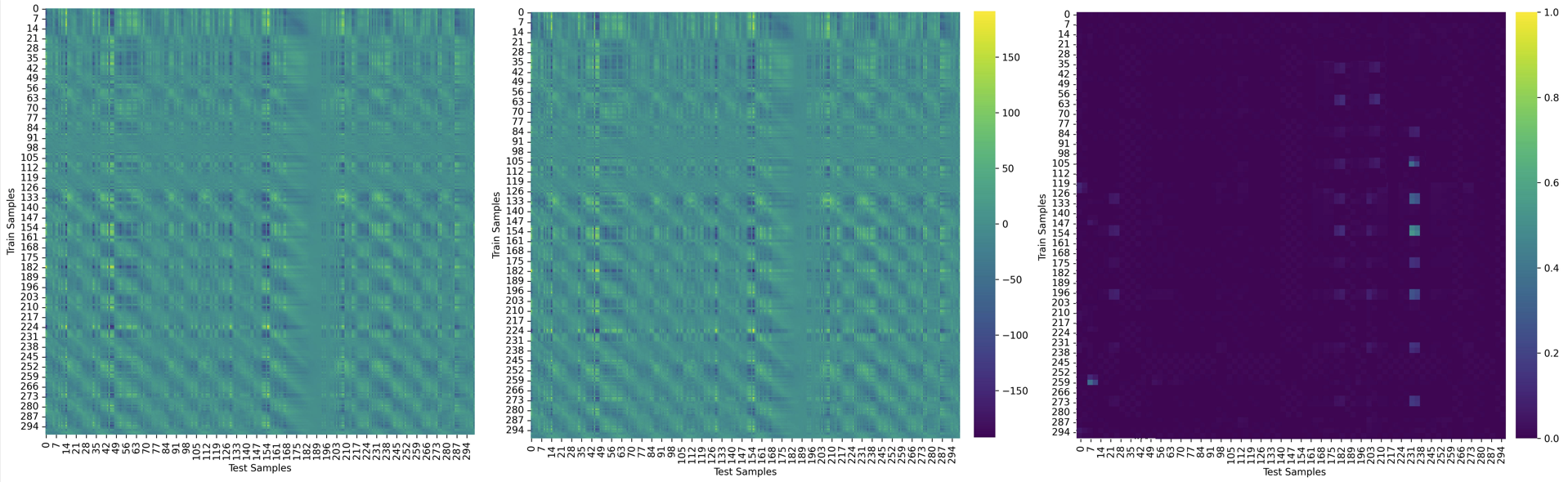}
        \caption{$\tau = 9$}
        \label{fig:k_sst2}
    \end{subfigure}

    \begin{subfigure}[b]{\textwidth}
        \includegraphics[width=0.95\textwidth]{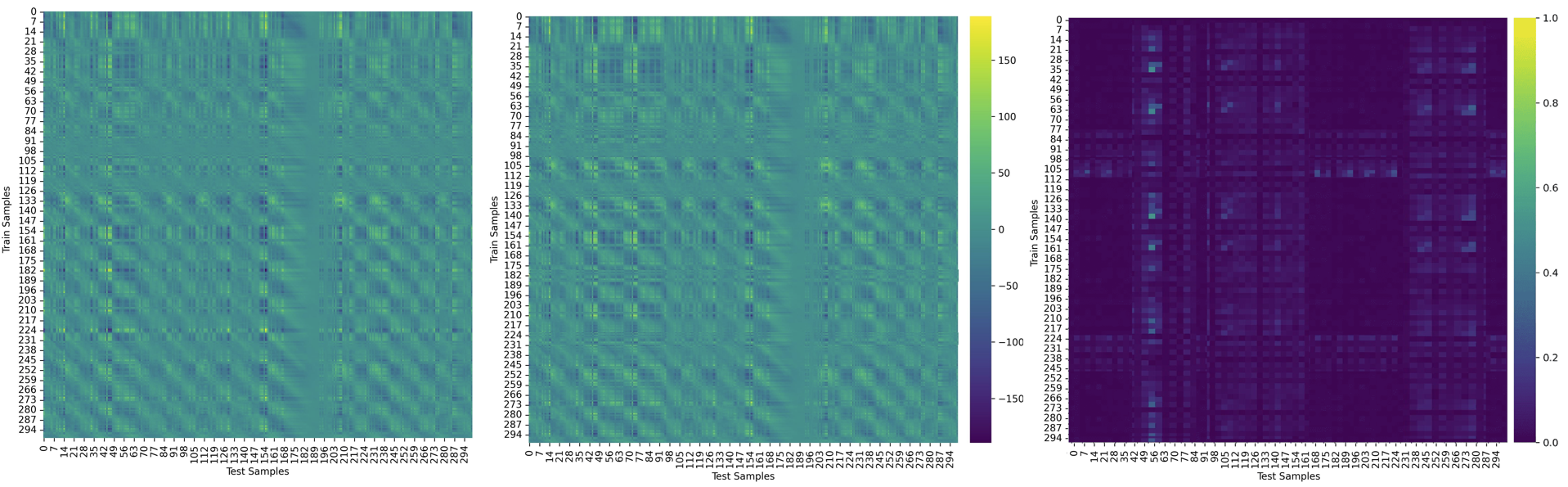}
        \caption{$\tau = 7$}
        \label{fig:k_rte}
    \end{subfigure}
    \begin{subfigure}[b]{\textwidth}
        \includegraphics[width=0.95\textwidth]{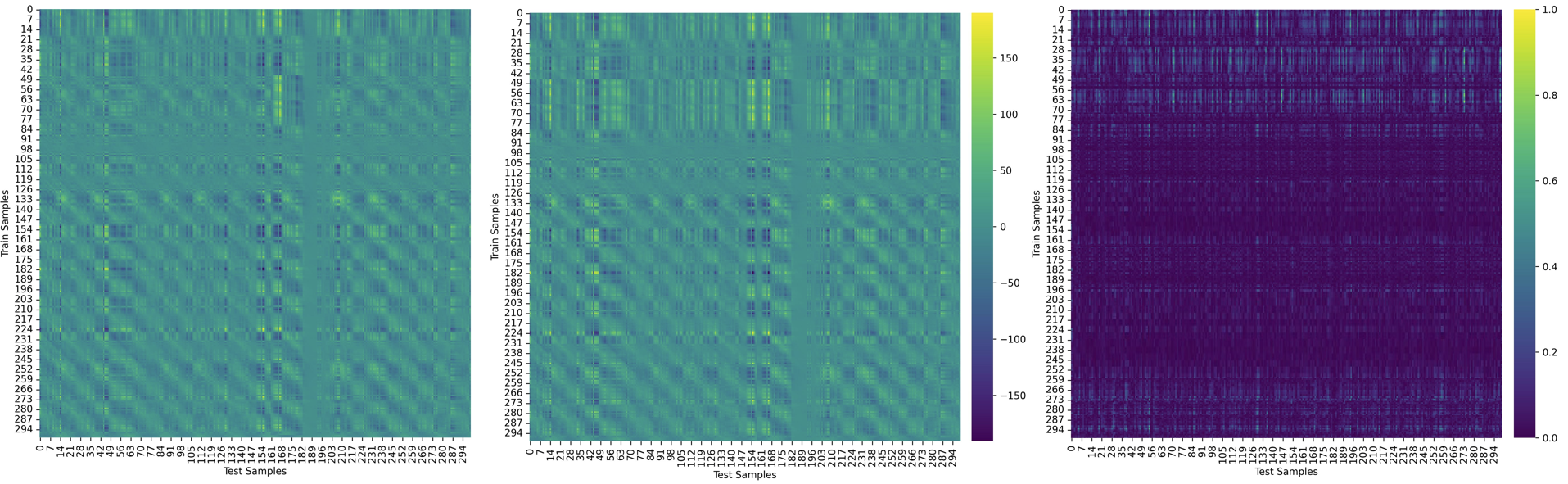}
        \caption{$\tau = 5$}
        \label{fig:k_cola}
    \end{subfigure}

    \caption{Visualization of (a) \(\tau = 9\), (b) \(\tau = 7\), and (c) \(\tau = 5\). Each subfigure displays the reference NTK at \(\tau = 10\), the NTK at the respective \(\tau\) level, and their absolute difference.}
    \label{fig:jacobian_heatmap}
\end{figure*}

\noindent\textbf{Short-term forecasting results: } In Table \ref{tab:full_short}, we present the short-term multivariate forecasting results, across four forecasting horizons: 12, 24, 48, and 96 time steps. Our proposed model consistently achieves low MSE and MAE values across the PEMS datasets, indicating a strong short-term predictive performance. Specifically, LLM-Mixer demonstrates competitive accuracy on PEMS03, PEMS04, and PEMS07, outperforming several baseline models, including TIME-LLM, TimeMixer, and PatchTST. Additionally, the LLM-Mixer shows robustness on PEMS08, where it delivers superior results compared to iTransformer and DLinear. These results emphasize the effectiveness of the LLM-Mixer in capturing essential temporal dynamics for short-horizon forecasting tasks.   

\noindent\textbf{Univariate forecasting results: } Table \ref{tab:full_uni} presents the univariate long forecasting results on the ETT benchmark and averaged over horizons of 96, 192, 384, and 720-time steps. LLM-Mixer achieves the lowest MSE and MAE values across all datasets, consistently outperforming other methods like Linear, NLinear, and FEDformer. LLM-Mixer demonstrates superior accuracy, particularly on most of the datasets. These results confirm the effectiveness of the LLM-Mixer in capturing complex temporal dependencies, solidifying its capability for univariate long-term forecasting.

\subsection{Ablation Study}
\textbf{Effect of Downsampling on Learning Dynamicse:}
To evaluate the impact of different downsampling levels on the learning dynamics of LLM-Mixer, we conducted an ablation study using the Neural Tangent Kernel (NTK) \cite{jacot2018neural}. Specifically, we aimed to understand how the number of downsampling levels affects the model's ability to capture multiscale information.
First, we used DeepEcho \cite{patki2016synthetic} to generate synthetic multivariate time series datasets for this study. We trained 10 versions of LLM-Mixer, each with a different number of downsampling levels \(\tau \in \{1, 2, \dots, 10\}\). For each model, we calculated the NTK on 300 sample pairs from both the training and test sets. The NTK, denoted as \(\mathbf{K}(\mathbf{x}, \mathbf{x}')\), is computed as the inner product of the gradients of the model outputs with respect to its parameters:
\[
\mathbf{K}(\mathbf{x}, \mathbf{x}') = \nabla_{\boldsymbol{\theta}} \theta_t(\mathbf{x}; \boldsymbol{\theta})^{\top} \nabla_{\boldsymbol{\theta}} \theta_t(\mathbf{x}'; \boldsymbol{\theta}),
\]
where \(\nabla_{\boldsymbol{\theta}} \theta_t(\mathbf{x}; \boldsymbol{\theta})\) is the gradient of the model output with respect to its parameters at iteration \(t\).

<span style="color: red;">Data Leakage Prevention Protocol: To ensure fair comparison and avoid data leakage, we construct prompts using only metadata and statistics computed exclusively from the training set. Specifically, we include: (1) dataset description (e.g., "electricity consumption data"), (2) feature names and units, (3) basic statistics (mean, standard deviation, data frequency) computed only from training samples. No information from validation or test sets is incorporated into the prompt construction process. We validate this approach through ablation studies comparing models with and without statistical information in prompts.</span>

 To measure how the NTK structure changes with different 10 levels, we used the Frobenius norm to calculate the distance between the NTK of each model (\(\mathbf{K}_{\tau}\)) and a reference NTK (\(\mathbf{K}_{10}\)), which corresponds to the model with the maximum downsampling levels. The NTK distance is defined as:
\[
 d_{\text{NTK}}(\tau) = \|\mathbf{K}_{10} - \mathbf{K}_{\tau}\|_F,
\]
where \(\|\cdot\|_F\) denotes the Frobenius norm. Smaller NTK distances indicate that the model's learning dynamics are closer to the reference model.

Our results, shown in Figure \ref{fig:ablation}, reveal that as the number of downsampling levels \(\tau\) decreases, the NTK distance increases. The largest distance is observed when \(\tau = 1\), indicating that using only one downsampling level significantly alters the model's learning dynamics.However, more downsampling levels are not always better. While increasing \(\tau\) enhances the model's ability to capture multiscale patterns, excessive downsampling may smooth out critical fine-grained details, which are essential for tasks with significant short-term variations. In Figure~\ref{fig:jacobian_heatmap}, we visualize the NTK of the reference model across different downsampling levels \(\tau\) and the normalized absolute differences.

\noindent\textbf{Multi-scale Mixing by Pooling: } We conducted an ablation study to explore the effects of various Multi-scale Mixing techniques. The techniques examined were Min, Max, Avg, and L2, each applying a unique method for aggregating downsampling information across scales.  Figure \ref{fig:ablation} (right) presents the MSE for each downsampling method across different datasets. Notably, average pooling consistently yielded a lower MSE, suggesting that this method is better suited for capturing multi-scale dependencies in the data.

\section{Conclusion}

This work introduces the LLM-Mixer, a novel framework that combines multiscale time-series decomposition with pre-trained LLMs for improved forecasting. By leveraging multiple temporal resolutions, the LLM-Mixer effectively captures both short-and long-term patterns, enhancing the model's predictive accuracy. Our experiments demonstrate that the LLM-Mixer achieves competitive performance across various datasets, outperforming recent state-of-the-art methods. 

\section{Limitations and Future Directions}

Although LLM-Mixer improves forecasting accuracy, several limitations warrant discussion. 

\textbf{Computational Requirements:} The use of pre-trained language models introduces significant computational overhead, which may limit deployment in real-time or resource-constrained environments. \textbf{Prompt Engineering:} Model performance depends on prompt quality and domain expertise for optimal prompt design, which may limit accessibility for non-experts. 

\textbf{Out-of-Distribution Robustness:} When training and test data distributions differ significantly, the fixed prompt approach may not adapt effectively to distributional shifts. 

\textbf{Limited Classical Baseline Analysis:} Our evaluation focuses primarily on deep learning methods and would benefit from comprehensive comparison with statistical approaches like ARIMA and exponential smoothing. 

\textbf{Data Leakage Potential:} While we implement protocols to prevent information leakage, the prompt-based approach requires careful validation to ensure fair comparison. 

\textbf{Domain Generalization:} Testing on more diverse domains (finance, healthcare, climate) would strengthen claims about broad applicability. Future work should address these limitations through adaptive prompting strategies, efficiency optimizations, and expanded empirical validation.

\bibliography{custom}

\appendix

% \section{Example Appendix}
\label{sec:appendix}

% This is an appendix.

\end{document}